\documentclass{article}
\usepackage{spconf,amsmath,epsfig}
\usepackage{multirow}

\let\OLDthebibliography\thebibliography
\renewcommand\thebibliography[1]{
  \OLDthebibliography{#1}
  \setlength{\parskip}{0pt}
  \setlength{\itemsep}{0pt plus 0.3ex}
}

\begin{document}\sloppy

% Example definitions.
% --------------------
\def\x{{\mathbf x}}
\def\L{{\cal L}}

% Title.
% ------
\title{SUMD: Super U-shaped Matrix Decomposition Convolutional neural network for Image denoising}
%
% Single address.
% ---------------
%Address and e-mail should NOT be added in the submission paper. They should be present only in the camera ready paper. 
\name{QiFan Li$^{\ast}$}
\address{$^{\ast}$l814794087@gmail.com}

\maketitle

\begin{abstract}
In this paper, we propose a novel and efficient CNN-based framework that leverages local and global context information for image denoising. Due to the limitations of convolution itself, the CNN-based method is generally unable to construct an effective and structured global feature representation, usually called the long-distance dependencies in the Transformer-based method. To tackle this problem, we introduce the matrix decomposition module(MD) in the network to establish the global context feature, comparable to the Transformer based method performance. Inspired by the design of multi-stage progressive restoration of U-shaped architecture, we further integrate the MD module into the multi-branches to acquire the relative global feature representation of the patch range at the current stage. Then, the stage input gradually rises to the overall scope and continuously improves the final feature. Experimental results on various image denoising datasets: SIDD, DND, and synthetic Gaussian noise datasets show that our model(SUMD) can produce comparable visual quality and accuracy results with Transformer-based methods.
\end{abstract}
\section{Introduction}
\label{sec:intro}

Image denoising is a fundamental and crucial task that is always used in the other high-level computer vision program as the front steps, receiving the attention of the academic circles. The main challenge in this task is to recover the clean image from the noisy image, and the problem can be modeled as:
\begin{eqnarray}
y &=& x+n
\end{eqnarray}
Where x, y, and n represent the clean image, noisy image, and noise matrix, respectively. Previous denoising methods can be divided into two mainstreams, and one is the traditional algorithm, which usually uses hypothetical prior information to address the problem, such as BM3D and NLM.

Following the development of deep learning, Methods based on the CNN and Transformer architecture use the amount of clean-noisy training pairs to acquire the implicit noise distribution and image priors for denoising. Recently, the Transformer-based methods have been used in many vision tasks and acquired notable performance due to advantages in long-distance dependent information. Combining the advantages of the CNN-based method in local representations and the Transformer-based methods in long-distance dependent information to restore high-quality and detailed images becomes a vital issue.

\begin{figure}
    \centering
    \includegraphics[width=1\linewidth]{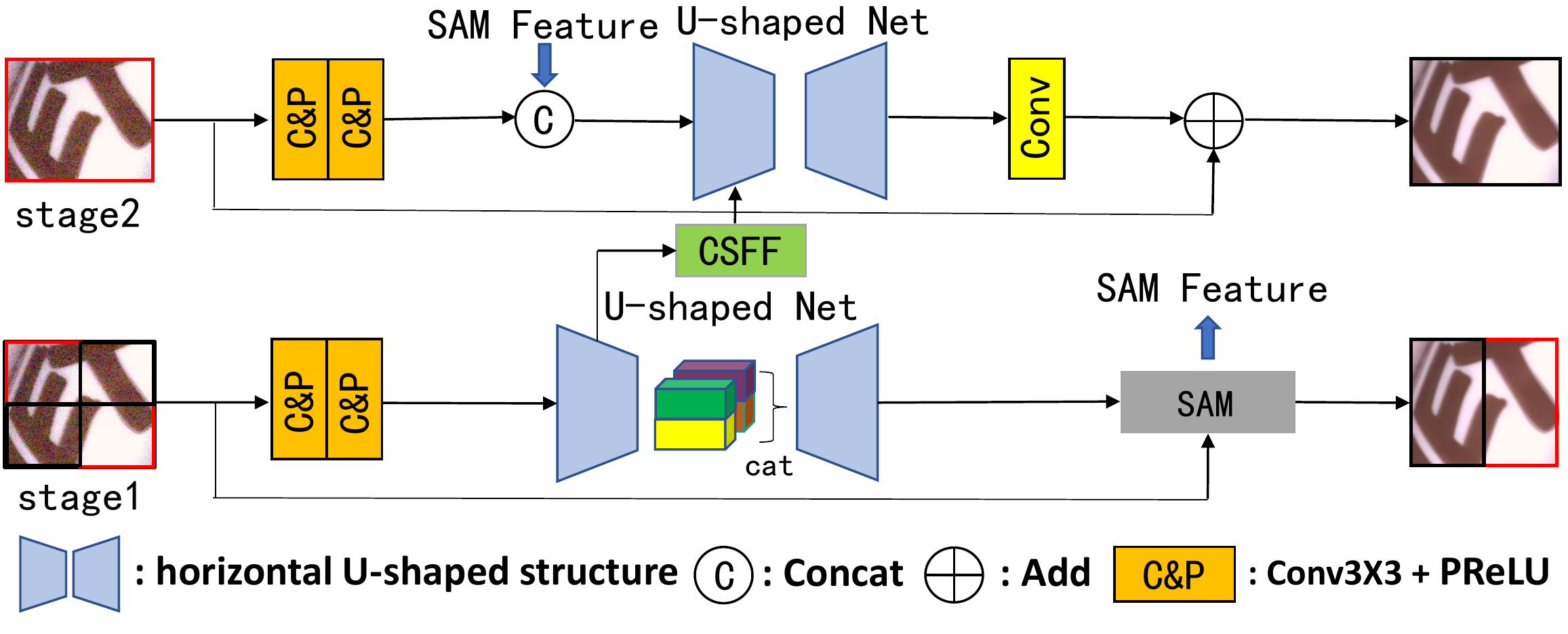}
    \caption{Overview of our SUMD architecture,containing two stages. We adopt CSFF and SAM modules from MPRNet~\cite{MPRNet}.}
    \label{fig:ST}
\end{figure}

This paper proposes SUMD, an efficient CNN-based architecture, to combine the advantages of the two methods mentioned in the previous paragraph for image denoising. We introduce the MD module to acquire the global context information to achieve these goals. Specifically, we reconstructed the low-rank information from the image to replace the long-distance dependent information extracted from the image. 

The MD module uses the non-negative matrix factorization method to reconstruct structural information, where the decomposition process uses the "one-step gradient" method to optimize named in ~\cite{MD}. At the same time, the parameters are initialized with a normal distribution and do not need to be saved that reduces a significant number of parameters. 

In order to further improve the low-rank global feature, we adopt the multi-stage progressive denoising strategy and build a relative global context feature within the current stage with different patch sizes through MD. Thereby the SUMD gradually denoises the more accurate images. Other denoising methods use the input patch to participate in the training process and add the image pyramid to obtain high-level information. The SUMD needs the high-level features and extracts the globally relative information using the non-overlapping patches as input. For example, their global feature representations are also different, with an image containing two independent objects(e.g., hands and T-shirt). It is evident that directly constituting the low-rank global representation will not pay attention to every detail for every object that appeared in the input and generate uncertain errors. Therefore, as mentioned earlier, while introducing multiple stages, the network at different stages is also focused on constructing global features at all levels in different scopes to achieve the goal of gradual recovery.

Each stage of our SUMD is built upon Unet~\cite{Unet} and keeps the same hierarchical encoder-decoder structure and the skip-connection with add MD module. It guarantees global context information and avoids the shortcomings of Transformer-based methods that cannot effectively use the local features. It should be noted that for low-level vision tasks, local features play a vital role in the final denoising of the image. Overall, we summarize the contributions of this paper included:
\begin{itemize}
    \item we propose SUMD, an effective and efficient multi-stage progressive network for image denoising.
    \item we assume the global context features as the low-rank information and utilize the matrix decomposition method to reconstruct it to further eliminate the interference of noise on the feature representation. The matrix decomposition approach is capable of acquiring the contextually-enriched and globally structured outputs.
    \item SUMD achieves state-of-the-art performance or comparable performance in both PSNR and SSIM indicators on many popular benchmarks.
\end{itemize}

\section{related work}
\textbf{Traditional Methods}
Traditional noise reduction algorithms have been a hot topic in the academic community in the past few decades. It has also promoted the development of methods based on deep learning in recent years. Previous work mainly used the prior hypothesis of the image or noise to design a model matching the hypothesis to perform the denoising program. Among them, the use of image prior includes sparse coding, non-local means, and the use of noise distribution assumptions includes AWGN denoiser. Although, these classical approaches have achieved good operating speed and accuracy results. However, the complexity and robustness of the traditional model are insufficient caused of the assumptions, and they cannot adapt to tasks in many situations.

\noindent
\textbf{Network architecture}
The deep learning-based model implicitly captures prior information from the training on the amount of clean-noisy data pairs for image denoising. Depending on their outstanding performance on the generalization, DNN based model used in varieties of the complex practical environment with a single training. With the various assumptions and novel ideas, MLP, TNRD, DnCNN, RBDN, FFDNet, MIRNet are also developed with promising denoising performance. MPRNet~\cite{MPRNet} relied on the multi-stage progressive recovery strategy, through supervised attention module, cross-stage feature fusion module, and the state-of-the-art performance had been achieved in the fields of denoising, deblurring, and de-raining. 

Recently, the Transformer architecture has gradually been applied to the field of computer vision and has made outstanding contributions to the global feature representation of long-range dependence. The Uformer~\cite{Uformer}, a transformer-based model, achieves the best performance in the field of image denoising. However, because relative position embedding still cannot replace the accurate relative position information, the extraction of local features is still poor.

\section{Method}
\begin{figure*}[t]
\centering
\begin{center}
\hfill
\begin{minipage}[b]{.32\linewidth}
  \centering
% \centerline{\epsfig{figure=image3.ps,width=4.0cm}}
  \includegraphics[width=1\linewidth]{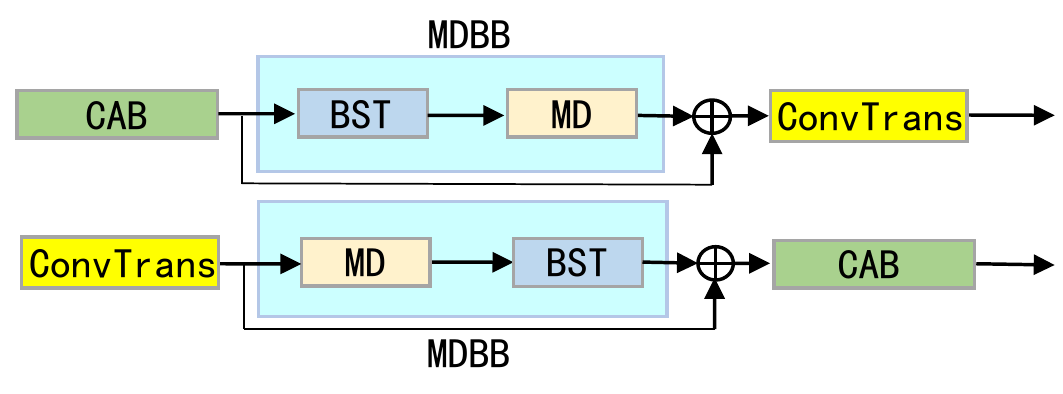}
  \centerline{(a)}\medskip
\end{minipage}
\begin{minipage}[b]{.27\linewidth}
  \centering
  \includegraphics[width=1\linewidth]{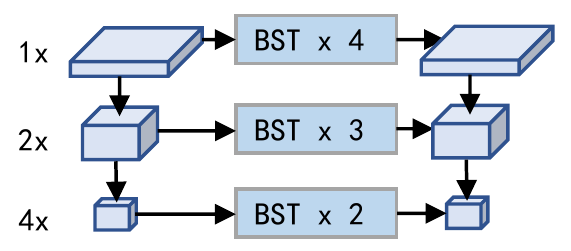}
  \centerline{(b)}\medskip
\end{minipage}
\begin{minipage}[b]{.18\linewidth}
  \centering
% \centerline{\epsfig{figure=image4.ps,width=4.0cm}}
  \includegraphics[width=.9\linewidth]{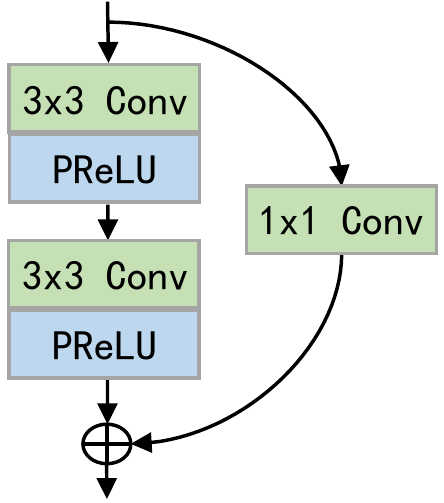}
  \centerline{(c)}\medskip
\end{minipage}
\begin{minipage}[b]{.19\linewidth}
  \centering
% \centerline{\epsfig{figure=image4.ps,width=4.0cm}}
  \includegraphics[width=.9\linewidth]{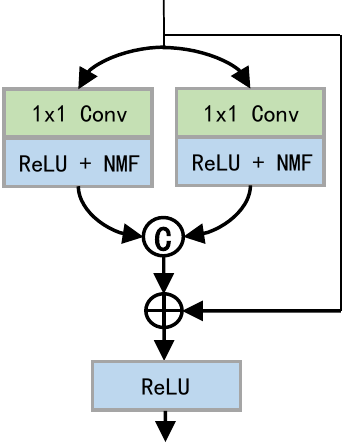}
  \centerline{(d)}\medskip
\end{minipage}
\end{center}
\caption{(a) Top: The encoder block. Bottom: The decoder block. (b) The skip-connection between the encoder block and the decoder block. (c) BST Module (d) MD Module}
\label{fig:components}
\end{figure*}

\begin{table*}[htpb]
\centering
\caption{The PSNR(dB) results of all competing methods on the three groups of test datasets. The best and second best results are highlighted in bold and Italic, respectively.} \label{tab:1}
\begin{tabular}{cccccccccccc}
\hline
\multirow{2}{*}{Case}  & \multirow{2}{*}{Datasets} & \multicolumn{8}{c}{Methods}                                                                                                                                                                                                           \\ \cline{3-11} 
                       &                           & \multicolumn{1}{c}{CBM3D} & \multicolumn{1}{c}{DnCNN-B} & \multicolumn{1}{c}{MemNet} & \multicolumn{1}{c}{FFDNet} & \multicolumn{1}{c}{UDNet} & \multicolumn{1}{c}{VDN}   & \multicolumn{1}{c}{NBNet} & \multicolumn{1}{c}{Uformer}         & Ours           \\ \hline
\multirow{3}{*}{Case1}  & Set5                      & \multicolumn{1}{c}{27.76} & \multicolumn{1}{c}{29.85}   & \multicolumn{1}{c}{30.10}  & \multicolumn{1}{c}{30.16}  & \multicolumn{1}{c}{28.13} & \multicolumn{1}{c}{30.39} & \multicolumn{1}{c}{\textit{30.59}} & \multicolumn{1}{c}{29.96} & \textbf{30.63} \\ \cline{2-11} 
                       & LIVE1                     & \multicolumn{1}{c}{26.58} & \multicolumn{1}{c}{28.81}   & \multicolumn{1}{c}{28.96}  & \multicolumn{1}{c}{28.99}  & \multicolumn{1}{c}{27.19} & \multicolumn{1}{c}{29.22} & \multicolumn{1}{c}{\textit{29.40}} & \multicolumn{1}{c}{29.36}  & \textbf{29.52} \\ \cline{2-11} 
                       & BSD68                     & \multicolumn{1}{c}{26.51} & \multicolumn{1}{c}{28.73}   & \multicolumn{1}{c}{28.74}  & \multicolumn{1}{c}{28.78}  & \multicolumn{1}{c}{27.13} & \multicolumn{1}{c}{29.02} & \multicolumn{1}{c}{\textit{29.16}} & \multicolumn{1}{c}{28.76} & \textbf{29.22} \\ \hline
\multirow{3}{*}{Case2}  & Set5                      & \multicolumn{1}{c}{26.34} & \multicolumn{1}{c}{29.04}   & \multicolumn{1}{c}{29.55}  & \multicolumn{1}{c}{29.60}  & \multicolumn{1}{c}{26.01} & \multicolumn{1}{c}{29.80} & \multicolumn{1}{c}{\textit{29.88}} & \multicolumn{1}{c}{29.53} & \textbf{30.11} \\ \cline{2-11} 
                       & LIVE1                     & \multicolumn{1}{c}{25.18} & \multicolumn{1}{c}{28.18}   & \multicolumn{1}{c}{28.56}  & \multicolumn{1}{c}{28.58}  & \multicolumn{1}{c}{25.25} & \multicolumn{1}{c}{28.82} & \multicolumn{1}{c}{\textit{29.01}} & \multicolumn{1}{c}{29.00} & \textbf{29.14} \\ \cline{2-11} 
                       & BSD68                     & \multicolumn{1}{c}{25.28} & \multicolumn{1}{c}{28.15}   & \multicolumn{1}{c}{28.36}  & \multicolumn{1}{c}{28.43}  & \multicolumn{1}{c}{25.13} & \multicolumn{1}{c}{28.67} & \multicolumn{1}{c}{\textit{28.76}} & \multicolumn{1}{c}{28.48} & \textbf{28.88} \\ \hline
\multirow{3}{*}{Case3} & Set5                      & \multicolumn{1}{c}{27.88} & \multicolumn{1}{c}{29.13}   & \multicolumn{1}{c}{29.51}  & \multicolumn{1}{c}{29.54}  & \multicolumn{1}{c}{27.54} & \multicolumn{1}{c}{29.74} & \multicolumn{1}{c}{\textit{29.89}} & \multicolumn{1}{c}{29.62}  & \textbf{30.01} \\ \cline{2-11} 
                       & LIVE1                     & \multicolumn{1}{c}{26.50} & \multicolumn{1}{c}{28.17}   & \multicolumn{1}{c}{28.37}  & \multicolumn{1}{c}{28.39}  & \multicolumn{1}{c}{26.48} & \multicolumn{1}{c}{28.65} & \multicolumn{1}{c}{\textit{28.82}} & \multicolumn{1}{c}{28.71} & \textbf{28.93} \\ \cline{2-11} 
                       & BSD68                     & \multicolumn{1}{c}{26.44} & \multicolumn{1}{c}{28.11}   & \multicolumn{1}{c}{28.20}  & \multicolumn{1}{c}{28.22}  & \multicolumn{1}{c}{26.44} & \multicolumn{1}{c}{28.46} & \multicolumn{1}{c}{\textit{28.59}} & \multicolumn{1}{c}{28.38} & \textbf{28.66} \\ \hline
\end{tabular}
\end{table*}

\subsection{Overview}

As shown in Fig~\ref{fig:ST}, the proposed framework is composed of two horizontal stages while each stage consists of the U-shaped encoder-decoder structure. The MD module is assimilated into the Encoder/Decoder block, and the basic design of the original Unet~\cite{Unet} architecture is modified to adapt to the requirement of denoising tasks. Then, the original noisy image is divided into multiple non-overlapping patches as the input in different stages. However, unlike the image pyramid, the input maintains the original resolution, which makes this stage pay more attention to the information within the patch. It facilitates the MD module to construct the relative global feature map. Between the two neighbor stages, we use the supervised attention module(SAM) and cross-stage feature fusion module(CSFF), where these two modules come from~\cite{MPRNet}. Specifically, given a noisy image $I \in R ^ {3 \times H \times W}$ at a certain stage, the SUMD extracts a low-level feature map $ F_{in} \in R ^ {C \times H \times W} $ through the initial block composed of a combination of a $ 3 \times 3 $ convolution and the PReLU stacking two blocks. Subsequently, the low-level features are delivered to the horizontal U-shaped structure to extract information from different scales gradually and then restored to the original resolution by degrees to obtain the decoded feature map $ F_{out} \in R ^ {C \times H \times W} $. The feature of the upper decoder in each stage is processed by the SAM module and then merged into the initial feature map of the next stage. The remaining encoder/decoder features of each scale will be adjusted by the CSFF module and incorporated into encoder/decoder features of the next stage. After the multi-stage processing, we apply a $ 3 \times 3 $ convolution on final decoder feature of the last stage to obtain a residual image $ F_{R} \in R ^ {3 \times H \times W} $. Finally, the denoising image is obtained by $I_{out} = I_{in}+F_{R}$.

\subsection{Horizontal U-shaped structure}

To meet the needs of the denoising task, we modified the components of the Unet~\cite{Unet} to boost the whole framework more efficient and simplified that illustrated in Fig~\ref{fig:components}(a). Above all, we utilize a channel attention block in~\cite{CABS} to weight the initial feature map $ F_{in} \in R ^ {C \times H \times W} $ with the channel self-attention mechanism for offering the channel information to promote the function of the MD module. Then, the weighted feature input to the MD module to extract the relative global low-rank structural features. The channel attention module and MD module form a basic encoder block and apply in the encoder part of each scale. Following the encoder block, we use a $4 \times 4$ convolution with stride 2 to down-sample the feature maps and increase the number of channels. For example, given the upper input feature $F_{0} \in R^{C_{in} \times H \times W}$, the $k$-th scale of the encoder produces the feature maps $F_{k} \in R^{(C_{in}+k \times C_{up})},k=0,1,2$. In order to effectively extract high-level context features, we set three scales for the image pyramid.

For the skip-connection between the encoder and the decoder at the same scale shown in Fig~\ref{fig:components}(b), we adopt stacked basic structure blocks(BST) shown in Fig~\ref{fig:components}(c) to further process features and maintain the details. For the image decoder, in addition to the features of the bottom scale that are direct as the input, the decoder block will not only receive the feature map refined by the skip-connection, but also the feature map of the previous scale will be up-sampled using $2 \times 2$ deconvolution with stride 2. The up-sampled feature decreases the channels and doubles the size of the feature maps. Unlike the encoder block, we switch the position of the channel attention block and the MD module, which stimulate the relative global feature selectively, to obtain better performance. 

\begin{figure*}[t]
\centering
\begin{center}
\hfill
\begin{minipage}[b]{.12\linewidth}
  \centering
  \includegraphics[width=.9\linewidth]{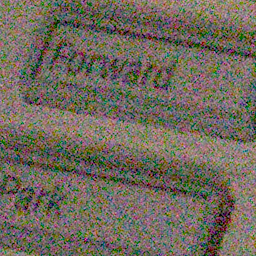}
  \centerline{18.25 dB}
  \centerline{Noisy}
\end{minipage}
\begin{minipage}[b]{.12\linewidth}
  \centering
% \centerline{\epsfig{figure=image4.ps,width=4.0cm}}
  \includegraphics[width=.9\linewidth]{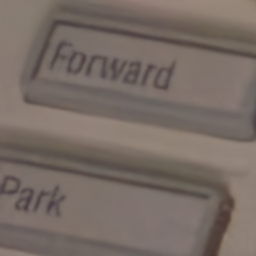}
  \centerline{36.39 dB}
  \centerline{VDN}
\end{minipage}
\begin{minipage}[b]{.12\linewidth}
  \centering
% \centerline{\epsfig{figure=image4.ps,width=4.0cm}}
  \includegraphics[width=.9\linewidth]{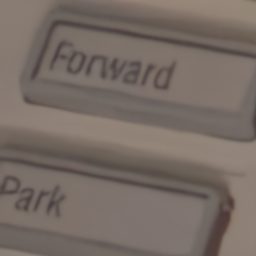}
  \centerline{36.72 dB}
  \centerline{CycleISP}
\end{minipage}
\begin{minipage}[b]{.12\linewidth}
  \centering
% \centerline{\epsfig{figure=image4.ps,width=4.0cm}}
  \includegraphics[width=.9\linewidth]{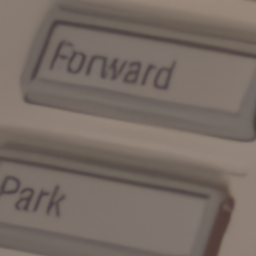}
  \centerline{36.98 dB}
  \centerline{MPRNet}
\end{minipage}
\begin{minipage}[b]{.12\linewidth}
  \centering
% \centerline{\epsfig{figure=image4.ps,width=4.0cm}}
  \includegraphics[width=.9\linewidth]{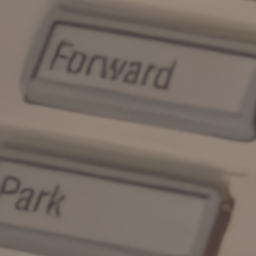}
  \centerline{36.97 dB}
  \centerline{MIRNet}
\end{minipage}
\begin{minipage}[b]{.12\linewidth}
  \centering
% \centerline{\epsfig{figure=image4.ps,width=4.0cm}}
  \includegraphics[width=.9\linewidth]{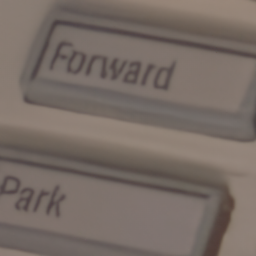}
  \centerline{36.89 dB}
  \centerline{Uformer}
\end{minipage}
\begin{minipage}[b]{.12\linewidth}
  \centering
% \centerline{\epsfig{figure=image4.ps,width=4.0cm}}
  \includegraphics[width=.9\linewidth]{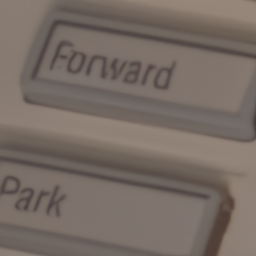}
  \centerline{37.11 dB}
  \centerline{Ours}
\end{minipage}
\begin{minipage}[b]{.12\linewidth}
  \centering
% \centerline{\epsfig{figure=image3.ps,width=4.0cm}}
  \includegraphics[width=.9\linewidth]{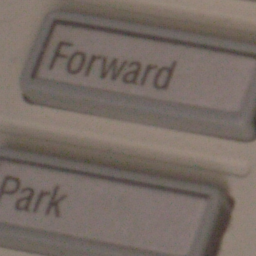}
  \centerline{}
  \centerline{Reference}
\end{minipage}
\quad
\begin{minipage}[b]{.12\linewidth}
  \centering
  \includegraphics[width=.9\linewidth]{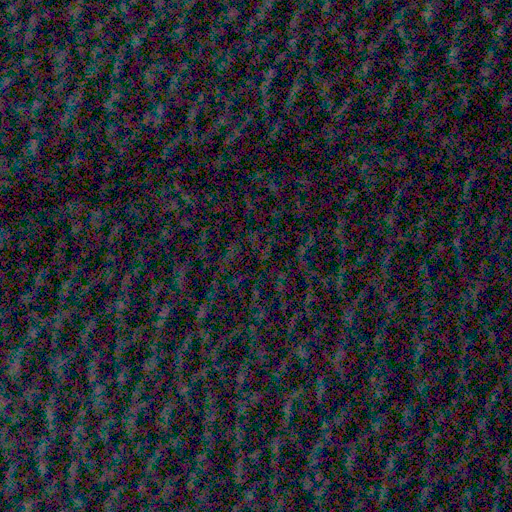}
  \centerline{}
  \centerline{Noisy}
\end{minipage}
\begin{minipage}[b]{.12\linewidth}
  \centering
  \includegraphics[width=.9\linewidth]{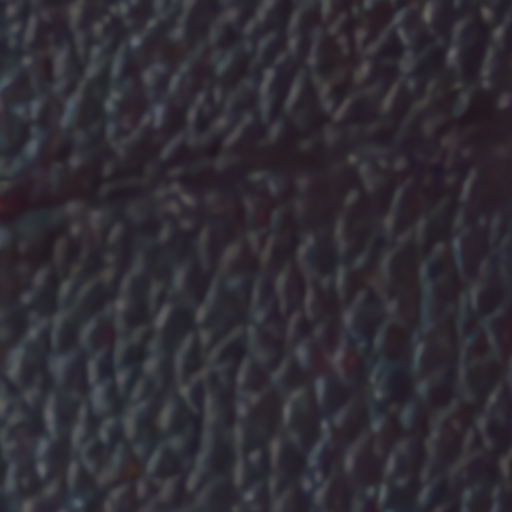}
  \centerline{32.09 dB}
  \centerline{VDN}
\end{minipage}
\begin{minipage}[b]{.12\linewidth}
  \centering
% \centerline{\epsfig{figure=image4.ps,width=4.0cm}}
  \includegraphics[width=.9\linewidth]{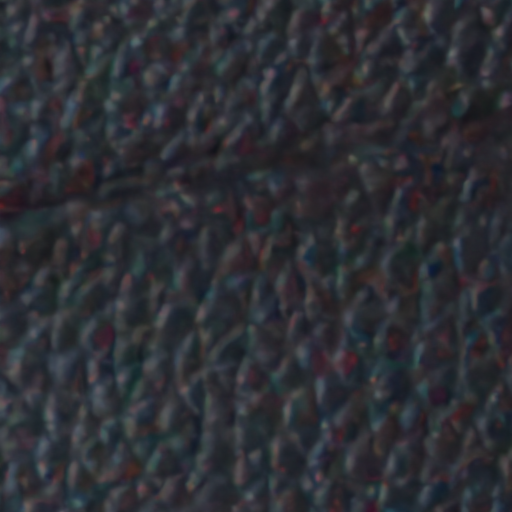}
  \centerline{32.27 dB}
  \centerline{CycleISP}
\end{minipage}
\begin{minipage}[b]{.12\linewidth}
  \centering
% \centerline{\epsfig{figure=image4.ps,width=4.0cm}}
  \includegraphics[width=.9\linewidth]{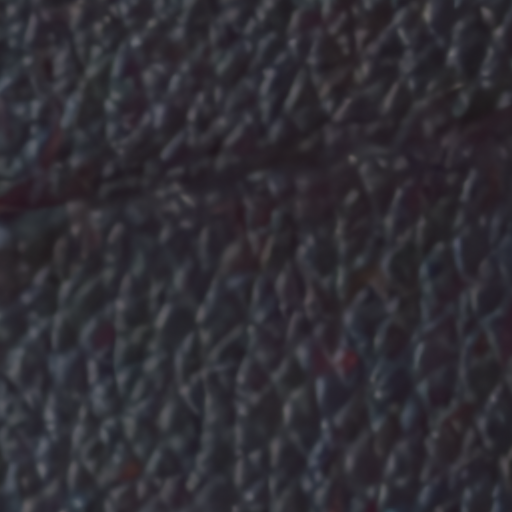}
  \centerline{32.63 dB}
  \centerline{MPRNet}
\end{minipage}
\begin{minipage}[b]{.12\linewidth}
  \centering
% \centerline{\epsfig{figure=image4.ps,width=4.0cm}}
  \includegraphics[width=.9\linewidth]{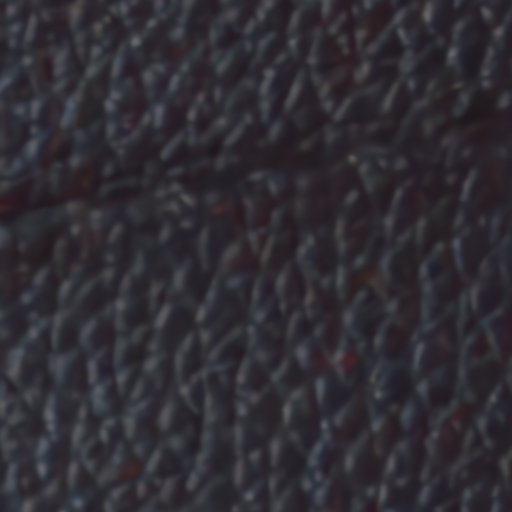}
  \centerline{32.54 dB}
  \centerline{MIRNet}
\end{minipage}
\begin{minipage}[b]{.12\linewidth}
  \centering
% \centerline{\epsfig{figure=image4.ps,width=4.0cm}}
  \includegraphics[width=.9\linewidth]{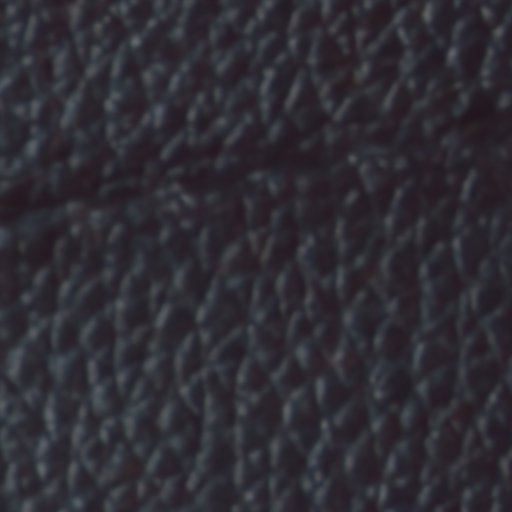}
  \centerline{33.27 dB}
  \centerline{NBNet}
\end{minipage}
\begin{minipage}[b]{.12\linewidth}
  \centering
% \centerline{\epsfig{figure=image4.ps,width=4.0cm}}
  \includegraphics[width=.9\linewidth]{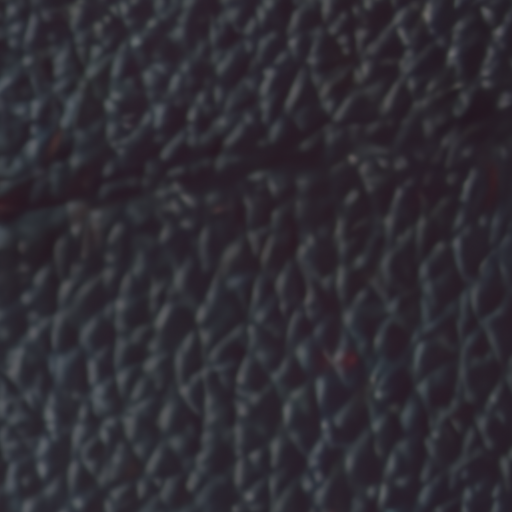}
  \centerline{33.04 dB}
  \centerline{Uformer}
\end{minipage}
\begin{minipage}[b]{.12\linewidth}
  \centering
% \centerline{\epsfig{figure=image4.ps,width=4.0cm}}
  \includegraphics[width=.9\linewidth]{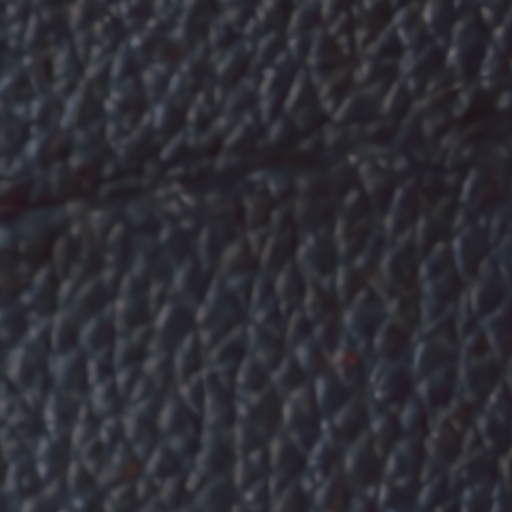}
  \centerline{33.70 dB}
  \centerline{Ours}
\end{minipage}

\end{center}
\caption{Visual results from SIDD and DND datasets. Top row: sample from the SIDD datasets. Bottom row: sample from the DND datasets. Our results preserve the details and reconstruct the textures.}
\label{fig:display}
\end{figure*}

\subsection{Matrix Decomposition Module}

Matrix decomposition is to factorize a matrix into the product of several sub-matrices. The helpful information is usually assumed as low-rank information for a noisy image and could be factorized into the low-rank matrix and the noise matrix using matrix decomposition. For the CNN-based denoising method, the convolution has pretty performance in capturing local features, but tough to construct the global feature maps limited by the convolution kernel size and receptive field. Therefore, the characteristic information obtained after matrix decomposition perfectly complements the convolution. The formula of the matrix decomposition can be expressed as:
\begin{eqnarray}
I_{n} &=&  I_{l} + N = DC + N
\end{eqnarray}
where $I_{n},I_{l},N \in R^{d \times hw}$ are the noisy image, the low-rank reconstruction and the noisy matrix, respectively. Among them, the factor of $I_l$ can be expressed as the product of the dictionary matrix $D \in R^{d \times r}$ and the corresponding codes $C\in R^{r\times hw}$,$h,w$ is the height and the width of input features.

The MD module is composed of preprocessing and matrix decomposition, shown in Fig~\ref{fig:components}(d). We use the BST to process the input features further to adapt to the matrix decomposition program as the pre-component. We firstly shallow the channel from the $F_{in}$ to $F_d$. Then, using a series of linear transformations consisting of the $1\times 1$ convolution and ReLU activation, we map the output of the preprocessing to split the $F_d$ into the $k$ subspaces of the features following the assumption that a union of multiple subspaces is hidden in $I_{l}$. After that, we utilize the Multiplicative Update rules~\cite{MUR} to solve the non-negative matrix factorization(NMF)~\cite{NMF}, decomposing and reconstructing the features. During the optimization step, we apply the one-step gradient to conquer the gradient vanishing, or gradient explosion described in~\cite{MD}. After that, we concat each decomposed feature and put it into another $1\times 1$ convolution to recover the original channels. The whole process can be expressed as:
\begin{align}
& F_k = W_kF_{in},    k=1,...,K  \nonumber \\
& D_k = MD(F_k),    k=1,...,K  \\
& F_{out}= W_{out}(Concat(D_1,...,D_k)),    k=1,...,K  \nonumber
\end{align}
where $F_{in} \in R^{C_{in} \times h \times w}$, $W_k \in R^{C_{mid} \times C_{in}}$ represents the input feature and the linear transformation for the $k$-th subspace, respectively. MD is the process of matrix decomposition. The input feature of the MD should be flattened along the H, W dimension. The $D_k \in R^{C_{mid} \times hw}$ of sub results are concated and then be transformed by the $W_{out} \in R^{C_{out} \times C_{mid \times K}}$. The $F_{out}$ is the reconstructed feature of the MD module.
\section{Experiments}
\begin{table*}[htpb]
\centering
\caption{The PSNR(dB) result of all competing methods on AWGN noise cases of three test datasets. The best and second best results are highlighted in bold and Italic, respectively.}
\label{tab:2}
\begin{tabular}{ccccccccccc}
\hline
\multirow{2}{*}{Case}     & \multirow{2}{*}{Datasets} & \multicolumn{9}{c}{Methods}                                                                                                                                                                                                           \\ \cline{3-11} 
                          &                           & \multicolumn{1}{c}{CBM3D} & \multicolumn{1}{c}{DnCNN-B} & \multicolumn{1}{c}{MemNet} & \multicolumn{1}{c}{FFDNet} & \multicolumn{1}{c}{UDNet} & \multicolumn{1}{c}{VDN}   & \multicolumn{1}{c}{NBNet}   & \multicolumn{1}{c}{Uformer}     & Ours           \\ \hline
\multirow{3}{*}{$\sigma$=15} & Set5                      & \multicolumn{1}{c}{33.42} & \multicolumn{1}{c}{34.04}   & \multicolumn{1}{c}{34.18}  & \multicolumn{1}{c}{34.30}  & \multicolumn{1}{c}{34.19} & \multicolumn{1}{c}{34.34} & \multicolumn{1}{c}{\textit{34.64}} & \multicolumn{1}{c}{34.32} & \textbf{34.74} \\ \cline{2-11} 
                          & LIVE1                     & \multicolumn{1}{c}{32.85} & \multicolumn{1}{c}{33.72}   & \multicolumn{1}{c}{33.84}  & \multicolumn{1}{c}{33.96}  & \multicolumn{1}{c}{33.74} & \multicolumn{1}{c}{33.94} & \multicolumn{1}{c}{\textit{34.25}} & \multicolumn{1}{c}{34.11} & \textbf{34.38} \\ \cline{2-11} 
                          & BSD68                     & \multicolumn{1}{c}{32.67} & \multicolumn{1}{c}{33.87}   & \multicolumn{1}{c}{33.76}  & \multicolumn{1}{c}{33.85}  & \multicolumn{1}{c}{33.76} & \multicolumn{1}{c}{33.90} & \multicolumn{1}{c}{\textit{34.15}} & \multicolumn{1}{c}{33.88} & \textbf{34.24} \\ \hline
\multirow{3}{*}{$\sigma$=25} & Set5                      & \multicolumn{1}{c}{30.92} & \multicolumn{1}{c}{31.88}   & \multicolumn{1}{c}{31.98}  & \multicolumn{1}{c}{32.10}  & \multicolumn{1}{c}{31.82} & \multicolumn{1}{c}{32.24} & \multicolumn{1}{c}{\textit{32.51}} & \multicolumn{1}{c}{32.13} & \textbf{32.60} \\ \cline{2-11} 
                          & LIVE1                     & \multicolumn{1}{c}{30.05} & \multicolumn{1}{c}{31.23}   & \multicolumn{1}{c}{31.26}  & \multicolumn{1}{c}{31.37}  & \multicolumn{1}{c}{31.09} & \multicolumn{1}{c}{31.50} & \multicolumn{1}{c}{\textit{31.73}} & \multicolumn{1}{c}{31.65} & \textbf{31.86} \\ \cline{2-11} 
                          & BSD68                     & \multicolumn{1}{c}{29.83} & \multicolumn{1}{c}{31.22}   & \multicolumn{1}{c}{31.17}  & \multicolumn{1}{c}{31.21}  & \multicolumn{1}{c}{31.02} & \multicolumn{1}{c}{31.35} & \multicolumn{1}{c}{\textit{31.54}} & \multicolumn{1}{c}{31.24} & \textbf{31.61} \\ \hline
\multirow{3}{*}{$\sigma$=50} & Set5                      & \multicolumn{1}{c}{28.16} & \multicolumn{1}{c}{28.95}   & \multicolumn{1}{c}{29.10}  & \multicolumn{1}{c}{29.25}  & \multicolumn{1}{c}{28.87} & \multicolumn{1}{c}{29.47} & \multicolumn{1}{c}{\textit{29.70}} & \multicolumn{1}{c}{28.98} & \textbf{29.74} \\ \cline{2-11} 
                          & LIVE1                     & \multicolumn{1}{c}{26.98} & \multicolumn{1}{c}{27.95}   & \multicolumn{1}{c}{27.99}  & \multicolumn{1}{c}{28.10}  & \multicolumn{1}{c}{27.82} & \multicolumn{1}{c}{28.36} & \multicolumn{1}{c}{\textit{28.55}} & \multicolumn{1}{c}{28.42} & \textbf{28.67} \\ \cline{2-11} 
                          & BSD68                     & \multicolumn{1}{c}{26.81} & \multicolumn{1}{c}{27.91}   & \multicolumn{1}{c}{27.91}  & \multicolumn{1}{c}{27.95}  & \multicolumn{1}{c}{27.76} & \multicolumn{1}{c}{28.19} & \multicolumn{1}{c}{\textit{28.35}} & \multicolumn{1}{c}{27.78} & \textbf{28.40} \\ \hline
\end{tabular}
\end{table*}

\begin{table*}[htpb]
\centering
\caption{Denoising results on the SIDD and DND datasets.}
\label{tab:3}
\begin{tabular}{cccccccccccc}
\hline
Datasets              & Method & BM3D  & CBDNet & RIDNet & VDN   & DANet & MPRNet & MIRNet & NBNet          & Uformer        & ours           \\ \hline
\multirow{2}{*}{SIDD} & PSNR   & 25.65 & 30.76  & 38.71  & 39.28 & 39.47 & 39.71  & 39.72  & 39.75          & \textit{39.77} & \textbf{39.77} \\ \cline{2-12} 
                      & SSIM   & 0.685 & 0.754  & 0.914  & 0.909 & 0.918 & 0.958  & 0.959  & 0.959          & \textit{0.959} & \textbf{0.959} \\ \hline
\multirow{2}{*}{DND}  & PSNR   & 34.51 & 38.06  & 39.26  & 39.38 & 39.59 & 39.80  & 39.88  & \textit{39.89} & \textbf{39.96} & 39.86          \\ \cline{2-12} 
                      & SSIM   & 0.851 & 0.942  & 0.953  & 0.952 & 0.955 & 0.954  & 0.956  & \textit{0.955} & \textbf{0.956} & 0.955          \\ \hline
\end{tabular}
\end{table*}

\subsection{Training details}

The proposed architecture adopts the end-to-end strategy during the training process and requires no extra component for pre-training. In the experiment, we use 2 stage horizontal U-shaped structures as a fundamental model and set the initial channel number as 128. After upsampling or downsampling in the image pyramid, the number of channels will increase or decrease by 48. In the MD module, we set the $F_{d}$ to 3/4 $F_{in}$, the number of subspace $k$ as 2, and $r=\sqrt{hw}/l,l=1,2,3$, $l$ is the level of the Horizontal U-shaped structure. We use the Charbonnier loss to train our model, the formula as :
\begin{equation}
\mathcal{L}(D,\textbf{\textit{x}},\textbf{\textit{y}})= \sqrt{\left\| D(\textbf{\textit{x}})-\textbf{\textit{y}} \right\| ^2 + \epsilon ^2},
\end{equation}
where $\textbf{\textit{x}},\textbf{\textit{y}},D(\cdot)$ represent noisy image, clean image and SUMD, respectively.

In the training phase, we use Adam optimizer with the momentum term (0.9,0.999) and the weight decay of $1e-8$. At the same time, set the initial learning rate to $2 \times 10^{-4}$ gradually reduced to $1 \times 10^{-6}$ with the cosine annealing. We crop the patches of size $128 \times 128$ from the training pairs as a sample with a batch size of 32 for 300K iterations. We augment the training sample with random horizontal flip and random rotation with $90^{\circ} \times k, k=1,2,3$.

\subsection{Synthetic Gaussian denosing}
Following the~\cite{VDN,nbnet}, we use the same experimental setup, in which the training samples are part of BSD, ImageNet validation datasets, Waterloo Exploration Database. The validation test datasets are composed of Set5, LIVE1, and BSD68. In the generation of the noise data, we are consistent with the method in~\cite{VDN,nbnet} to facilitate fair comparison. We generate four noise masks, one of which is used for training and the other three types used for testing. The specific non-$i.i.d.$ Gaussian noise generation formula is as follows:
\begin{eqnarray}
\textbf{\textit{n}} &=& \textbf{\textit{n}}^1 \odot \textbf{\textit{M}} ,  n^1_{ij} \sim \mathcal{N}(0,1),
\end{eqnarray}
where $\textbf{\textit{M}}$ is a spatially variant mask. We can demonstrate the denoising results for multiple types of synthetic Gaussian noise under the training of a category of noise data. 

Table~\ref{tab:1} lists the denoising performance of 4 types of non-$i.i.d.$ Gaussian noise and compares them with the previous methods. From the table~\ref{tab:1}, our algorithm exceeds the previous arithmetic VDN~\cite{VDN} and NBNet~\cite{nbnet} and acquires an average improvement of more than 0.1 dB on the PSNR compared with the NBNet in every test case. Table~\ref{tab:2} illustrates the PSNR results in the scene of additive white Gaussian noise(AWGN) and accomplishes the better performance(average increase of 0.1 dB).

Our method uses the MD module to reconstruct the context features without prior noise data distribution gradually. On the one hand, it separates the signal and the noise. On the other hand, it also makes the recovery of the detailed information of the image more accurate within the relative ranges.

\subsection{Real Image denoising}
\textbf{SIDD Dataset}  To justify that our model is competent for real noise scenarios, we use the SIDD~\cite{SIDD}, a commonly used mobile phone real-noise benchmark, to evaluate the performance. We use 320 high-resolution images to train our model and evaluate the performance of the 1280 validation patches offered by the SIDD. We compare with the previous methods, such as VDN~\cite{VDN}, MPRNet~\cite{MPRNet}, MIRNet, NBNet~\cite{nbnet} and Uformer\cite{Uformer}, and quantitatively demonstrated the denoising capacity of various models in Table 3. We achieve the comparable performance of 39.77 dB with Uformer on PSNR and surpass the other methods at least 0.02 dB. In addition, we also provide a visual denoising comparison example from different models in the top row of Fig~\ref{fig:display}.

\noindent
\textbf{DND Dataset} We further utilize the DND~\cite{DND} datasets to demonstrate the capability of our model. Unlike the NBNet training strategy, we only train the model on the SIDD training set and directly use the best model from the SIDD, evaluating the performance on the DND benchmark that contained 1000 patches for testing. We present the visual comparisons of our results with the other state-of-the-art methods in Fig~\ref{fig:display}, and our model obtains 39.86 dB shown in Table~\ref{tab:3}, the comparable results without using any extra datasets. For instance, our method achieves 0.66 dB improvement over Uformer on DND.

\begin{table}[htpb]
\centering
\caption{Ablation study on MD and other modules of the proposed in SUMD.}
\label{tab:4}
\begin{tabular}{ccc}
\hline
\#Stages & Stage Combination & PSNR  \\ \hline
2        & Unet              & 39.56 \\
2        & Unet+HUS          & 39.73 \\
2        & Unet+HUS+MD          & 39.77 \\ \hline
1        & Unet+HUS+MD       & 39.68   \\ \hline
\end{tabular}
\end{table}
\begin{table}[htpb]
\centering
\caption{Effect of the middle dimension of the MD module. The rate represents the original dimensions proportioned to the middle dimensions.}
\label{tab:5}
\begin{tabular}{cccccc}
\hline
ratio & w/o   & 1:4   & 2:4   & 3:4   & 4:4   \\ \hline
PSNR  & 39.49 & 39.55 & 39.58 & 39.60 & 39.55 \\ \hline 
\end{tabular}
\end{table}

\subsection{Ablation Studies}
\textbf{Multi-stages with MD Module} To demonstrate the performance of our MD module and HUS structure, we test each component on the various network settings. Table~\ref{tab:4} shows that the HUS merged into Unet has improved the 0.17 dB higher than the simple Unet structure in 2 stages on PSNR. Then, we compare our regular version to the model that replaced the MD module with a BST to keep a fair comparison. The results demonstrate a substantial drop in PSNR from 39.77 dB to 39.73 dB when we remove the MD module. From Table~\ref{tab:4}, we can also justify the effectiveness of the multi-stages architecture in our model. Compared with the single-stage and the two stages design.  The PSNR has improved from 39.68 dB to 39.77 dB.

\noindent
\textbf{Integrated into simple Unet with HUS} In order to examine the performance of the design on a simple CNN denoising model, We shallow the channels from 128 to 80 and test the model with different middle dimensions on a single-stage Unet with HUS structure. In Table~\ref{tab:5}, We notice that the CNN with MD module can achieve about 0.1 dB higher than the w/o MD module. 

\noindent
\textbf{Influence about the reducing rate} In Table~\ref{tab:5}, We also explore the appropriate K, a reducing rate of the middle dimensions of the MD module, boosting our model achieving the best performance. It is evident that when we set the rate k between 1/2 and 3/4, the MD can realize a low-ranked, compact, and representative feature to tackle the shortcoming of the CNN-based model that lacks global information.

\section{Conclusion}
In this work, we present a multi-stage progressive CNN-based model with the MD module to tackle the problem of global feature extraction. To compare with the Transformer long-distance dependent feature, we assume that the low-ranked features can substitute the global features and achieve comparable performance. In order to boost the MD module, we also adopt the multi-stages design that can limit the receptive field in a definite scope on the current stage and the channel attention block to weight feature from the different channels for promoting the critical feature maps decided in the final process. We also believe that how to utilize the MD module to extract the global feature will be a vital issue that can facilitate the CNN model to acquire the global structured features.
\bibliographystyle{IEEEbib}
\bibliography{icme2022template}

\end{document}